\documentclass{article}

\usepackage{microtype}
\usepackage{graphicx}
\usepackage{subcaption}
\usepackage{booktabs}
\usepackage{longtable}
\usepackage{array}
\usepackage[most]{tcolorbox}
\usepackage{enumitem}
\usepackage{listings}

\usepackage{hyperref}
\usepackage{xurl}

\usepackage[accepted]{icml2026}

\usepackage{amsmath}
\usepackage{amssymb}
\usepackage{mathtools}
\usepackage{amsthm}
\usepackage[capitalize,noabbrev]{cleveref}

\makeatletter
\ifdefined\isaccepted\else
  \ifdefined\ispreprint\else
    \renewcommand{\printAffiliationsAndNotice}[1]{%
      \global\icml@noticeprintedtrue%
      {\let\thefootnote\relax\footnotetext{\hspace*{-\footnotesep}\Notice@String}}%
    }
  \fi
\fi
\makeatother

\newcommand{\ActionToken}[1]{\ifmmode\mathtt{#1}\else\texttt{#1}\fi}
\newcommand{\Aone}{\ActionToken{Ax1}}
\newcommand{\Atwo}{\ActionToken{Ax2}}
\newcommand{\Athree}{\ActionToken{Ax3}}

\newcommand{\MP}{\ActionToken{MP}}
\newcommand{\LibAct}[1]{\ActionToken{Thm#1}}
\newsavebox{\fitcellbox}
\newcommand{\fitcell}[1]{%
  \sbox{\fitcellbox}{\fontsize{6.7}{6.7}\selectfont\scalebox{0.96}{$#1$}}%
  \ifdim\wd\fitcellbox>\linewidth
    \resizebox{\linewidth}{!}{\usebox{\fitcellbox}}%
  \else
    \usebox{\fitcellbox}%
  \fi
}

\graphicspath{{figures/}}

\definecolor{contributionblue}{HTML}{4594c1}
\newcommand{\contribution}[1]{%
\begin{tcolorbox}[
  enhanced,
  colback=contributionblue!8!white,
  colframe=contributionblue,
  leftrule=1.2mm,
  rightrule=0mm,
  toprule=0mm,
  bottomrule=0mm,
  arc=0mm,
  boxsep=0pt,
  left=4pt,
  right=4pt,
  top=2pt,
  bottom=2pt,
  breakable
]
#1
\end{tcolorbox}
}

\icmltitlerunning{Self-Supervised Theorem Discovery in a Formal Axiomatic System}

\begin{document}

\twocolumn[
  \icmltitle{Self-Supervised Theorem Discovery in a Formal Axiomatic System}

  \begin{icmlauthorlist}
    \icmlauthor{Kazuki Ota}{utokyo,riken}
    \icmlauthor{Takayuki Osa}{riken}
    \icmlauthor{Tatsuya Harada}{utokyo,riken}
  \end{icmlauthorlist}

  \icmlaffiliation{utokyo}{The University of Tokyo, Japan}
  \icmlaffiliation{riken}{RIKEN AIP, Japan}

  \icmlcorrespondingauthor{Kazuki Ota}{ota@mi.t.u-tokyo.ac.jp}

  \icmlkeywords{Machine Learning, Automated Theorem Proving, Mathematical Reasoning}

  \vskip 0.3in
]

\printAffiliationsAndNotice{}

\begin{abstract}
Recent artificial intelligence (AI) systems have shown remarkable progress in mathematical reasoning.
Many existing approaches, including large language models (LLMs), draw on human prior knowledge in the form of mathematical text, code, or theorem libraries.
Although these approaches are highly effective in practice, it remains an open question whether an agent can autonomously discover useful theorems without such human priors.
We study this question in a formal axiomatic system by developing an agent that starts from axioms and inference rules alone and gradually grows a library of useful theorems.
Concretely, we propose a self-supervised theorem-discovery algorithm that alternates between proof search and useful-theorem extraction, building a theorem library whose entries are reused as lemmas for subsequent proof search.
Experiments show that the agent discovers tens of thousands of theorems and finds proofs for human-written benchmark problems, suggesting that its discoveries include theorems meaningful from a human mathematical perspective.
Furthermore, the discovered theorems improve LLM proof performance when provided as prompt lemmas, indicating that they can serve as external knowledge for LLM reasoning.
Our results provide evidence that useful theorems can emerge from proof search without relying on human-provided theorem libraries.
More broadly, they suggest a path toward self-evolving AI systems for mathematics whose discoveries remain formally verifiable.
\end{abstract}

\section{Introduction}

Artificial intelligence (AI) systems are rapidly improving their mathematical reasoning abilities.
Large language models (LLMs) such as GPT and Gemini have shown steady gains on mathematical benchmarks \citep{OpenAI2024o1,OpenAI2025GPT5SystemCard,GeminiTeam2025Gemini25}.
A recent Gemini model was also reported to achieve gold-medal standard performance at the International Mathematical Olympiad 2025 \citep{LuongLockhart2025GeminiIMO}, highlighting advances in LLM-based mathematical reasoning.

Although LLMs can perform mathematical reasoning in natural language, their outputs can be difficult to verify because they may contain plausible but incorrect statements caused by hallucination \citep{Farquhar2024SemanticEntropy,Kalai2025Hallucination}.
This has motivated work on theorem proving with formal languages and proof assistants as a path toward rigorous mathematical reasoning.
Recent work includes autoformalization and proof sketching methods that connect natural-language mathematics to formal proofs \citep{Wu2022Autoformalization,Jiang2023DSP}, methods for generating proofs in proof assistants such as Lean \citep{Han2022PACT,Polu2022StatementCurriculum,Yang2023LeanDojo,Xin2024DeepSeekProver,Zimmer2025Bourbaki,Hubert2026AlphaProof}, and methods that combine language-model generation with symbolic reasoning or learned libraries \citep{Trinh2024AlphaGeometry,Wang2024LEGOProver,Dong2025STP}.
This line of work highlights the value of combining formally verifiable reasoning substrates with the search capabilities of machine learning models.

At the same time, many current methods start from mathematical knowledge constructed by humans.
For example, mathematical language models may use natural-language mathematical knowledge through pretraining on mathematical text and code \citep{Azerbayev2024Llemma}, and methods that connect natural language to formal proofs often start from human-written problem statements or informal proofs \citep{Wu2022Autoformalization,Jiang2023DSP}.
In addition, methods for proof generation and retrieval-augmented theorem proving often use existing formal libraries and the theorems and proofs contained in them \citep{Han2022PACT,Polu2022StatementCurriculum,Yang2023LeanDojo}.
Even methods based on large-scale synthetic data often construct training data from natural-language mathematical problems \citep{Xin2024DeepSeekProver}.
Such knowledge is highly useful in practice, but it makes it difficult to isolate the question of how much mathematical structure an agent can discover from formal rules alone.

A related line reduces reliance on human-provided proofs or existing data by learning from axioms and formal rules \citep{Wu2021TacticZero,Laurent2022RefineSearch,Poesia2024Minimo}.
These studies show that agents can self-improve in axiomatized domains and discover theorems or proofs.
\citet{Kasriel2025Usefulness} is especially close, as it discovers theorems from axioms, reuses them for further search, and evaluates external usefulness by asking an LLM judge whether they appear useful.
However, it remains open whether such theorems can improve an external reasoning agent's proof performance when supplied as lemmas.
We address this question and show that the theorems discovered by our approach improve both the agent's own proof search and LLM proof search.

In this study, we ask whether an agent can autonomously discover useful theorems without relying on human prior knowledge.
To isolate this question, we require the agent to start only from the primitive rules of a formal axiomatic system, without using an existing theorem library, a proof corpus, natural-language priors, or an externally provided training problem set.
Human-written problems are used only for evaluation, not for training the agent or constructing the theorem library.

We instantiate this formal-system setting in a Hilbert axiom system for propositional logic.
This setting is deliberately minimal: it has only a few primitive rules, but proof search is still nontrivial and every discovered theorem remains formally checkable.
Specifically, we formulate Hilbert-system proof search as a stack-machine decision process, where the state is the current proof stack and the actions push Hilbert axioms onto the stack or apply Modus Ponens.
Because formulas obtained during search can be regarded as theorems, proof search itself generates theorem candidates.

On top of this stack-machine formulation, we propose a self-supervised theorem-discovery algorithm that grows a theorem library from the agent's own proof search.
The agent reuses theorems reached during search as future proof goals and learns a goal-conditioned policy from the action sequences that reach them.
The supervision comes from the agent's own successful proof prefixes: whenever search reaches a theorem, the corresponding action sequence becomes a training signal for proving that theorem again.
It then extracts useful theorems from the discovered set and adds them as theorem actions that can be used as lemmas in later generations of proof search.
In this way, the agent learns not only to discover theorems, but also to use them to prove new theorems.

Experiments show that the proposed method discovers theorems that are useful both for the agent's own proof search and for proof search by external LLMs.
Specifically, the agent discovers tens of thousands of theorems and finds proofs for human-written benchmark problems.
Moreover, the extracted theorems improve LLM proof performance when provided as prompt lemmas.
These results show that useful theorems can emerge from proof search based only on axioms and inference rules, and that they can serve not only as internal lemmas for the agent but also as knowledge that assists external reasoning systems.

Our contributions can be summarized as follows:
\contribution{
\begin{enumerate}[leftmargin=1.35em,labelsep=0.4em,itemsep=0.2ex,topsep=0.2ex,parsep=0pt,partopsep=0pt]
\item We formulate propositional theorem proving as a stack-machine decision process based directly on the axioms and the inference rule of a Hilbert axiom system.
\item We propose a self-supervised theorem-discovery algorithm that grows a theorem library from the primitive rules of a Hilbert axiom system and then reuses extracted theorems as lemmas for further proof search.
\item Our experiments show that our agent discovers tens of thousands of theorems and finds proofs for human-written benchmark problems.
\item Finally, we show that the extracted theorems also improve LLM proof performance when provided as prompt lemmas, serving as external knowledge for LLM reasoning.
\end{enumerate}
}

\section{Preliminaries}
\label{sec:preliminaries}

\begin{figure*}[t]
\centering
\includegraphics[width=0.95\textwidth]{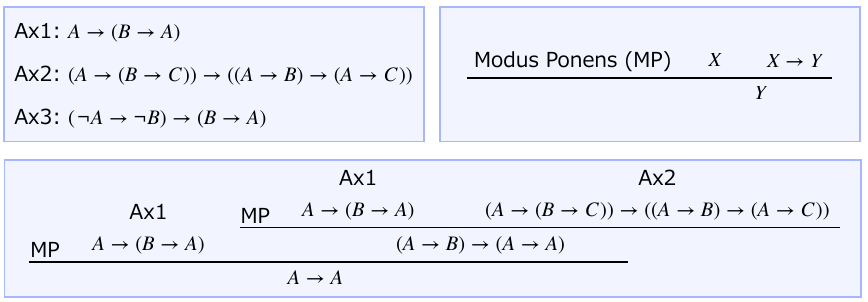}
\caption{Hilbert system and proof tree for $A \to A$. Top: the three axioms and Modus Ponens. The axioms respectively mean weakening, distribution of implication, and contraposition. Bottom: a proof tree deriving $A \to A$ from two \Aone{} instances, one \Atwo{} instance, and two Modus Ponens steps.}
\label{fig:hilbert-system}
\end{figure*}

This section defines the logical setting used throughout the paper.
It provides the basis for the stack-machine decision process introduced in Section~\ref{sec:stack-machine-formulation}.

\paragraph{Notation for Propositional Formulas.}
To keep syntax simple, this paper builds formulas only from implication $\to$ and falsity $\bot$.
Standard presentations often use primitive connectives such as negation $\neg$, conjunction $\land$, disjunction $\lor$, implication $\to$, and biconditional $\leftrightarrow$.
For expressive power in classical propositional logic, these connectives need not all be primitive, since implication and falsity form a basis for the usual connectives \citep{Post1921}.
Concretely, define $\neg A := A \to \bot$, $A \lor B := (A \to \bot) \to B$, and $A \land B := (A \to (B \to \bot)) \to \bot$.
Define $A \leftrightarrow B$ as $(A \to B) \land (B \to A)$, expanded using conjunction.
Thus, unless otherwise stated, every formula uses only $\to$ and $\bot$, and other connectives are syntactic sugar expanded by the translations above.

\paragraph{Rules of the Hilbert System.}
A Hilbert system is an axiomatic proof system in which theorems are derived from axioms using inference rules \citep{HilbertAckermann1950,Mendelson2015}.
The system used in this paper, shown in Figure~\ref{fig:hilbert-system}, consists of three axioms, \Aone, \Atwo, \Athree, and one inference rule, Modus Ponens.
The symbols $A,B,C$ in the axioms stand for arbitrary formulas.
A proof repeatedly introduces axioms and applies Modus Ponens, which derives $Y$ from previously obtained formulas $X$ and $X \to Y$.
When the final formula matches the target, it has been derived as a theorem.
The bottom panel of Figure~\ref{fig:hilbert-system} shows a proof tree for the theorem $A \to A$.
It uses two instances of \Aone, one instance of \Atwo, and two applications of Modus Ponens.
In this way, a Hilbert system proves theorems by combining axioms and inference rules.

\paragraph{Soundness and Completeness.}
This Hilbert system is known to be sound and complete for classical propositional logic \citep{Church1996}.
Soundness means that every formula derived in this system is valid under the standard semantics of classical propositional logic.
Completeness gives the converse guarantee, namely every formula valid in classical propositional logic can be derived in this Hilbert system.
These properties justify viewing the Hilbert system as one of the standard axiomatic systems for classical propositional logic.

\paragraph{Difficulty of Finding Proofs.}
While soundness and completeness guarantee correctness and existence of proofs, they do not imply that proofs are easy to find.
Indeed, deciding propositional validity is a coNP-complete problem \citep{Cook1971}.
Moreover, from the perspective of proof complexity, valid formulas need not have short proofs in a given proof system, and proof length can vary substantially across proof systems \citep{CookReckhow1979}.
Therefore, efficiently finding proofs in the Hilbert system is a nontrivial problem.

\section{Stack-Machine Formulation of Theorem Proving}
\label{sec:stack-machine-formulation}

\begin{figure*}[t]
\centering
\includegraphics[width=0.95\textwidth]{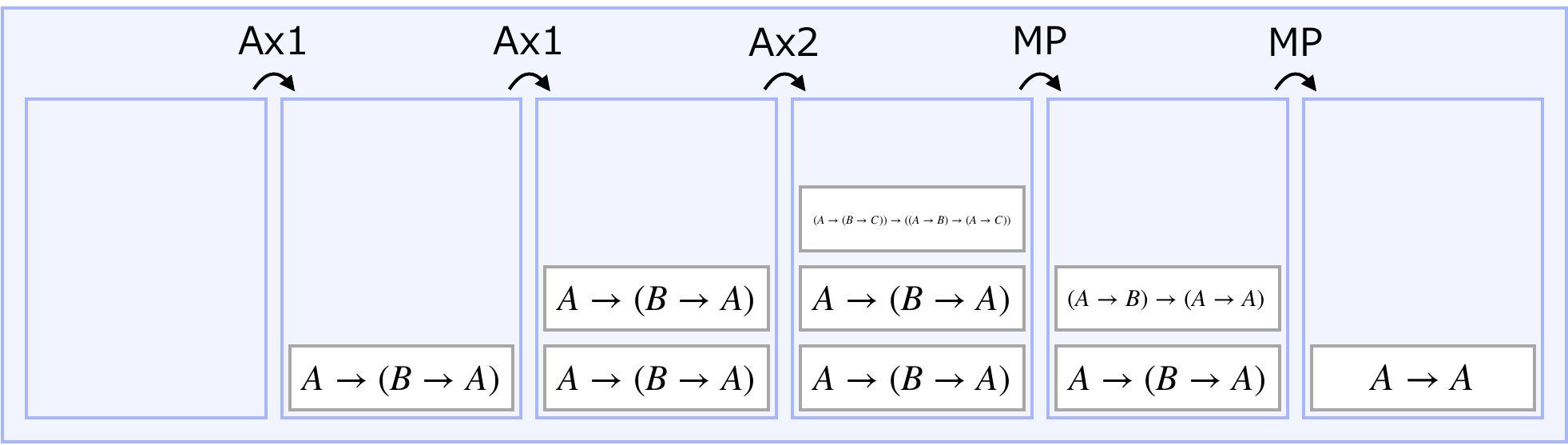}
\caption{%
Stack-machine execution for the proof tree of $A \to A$.
The proof tree in Figure~\ref{fig:hilbert-system} is converted into a linear action sequence: axiom instances at the leaves are executed as push actions \Aone, \Atwo, and \Athree{}, and internal Modus Ponens nodes are executed as \MP{} actions.
Each \MP{} action pops an implication and its antecedent from the stack and pushes the corresponding consequent.
This example shows that the five-action sequence \Aone, \Aone, \Atwo, \MP, \MP{} is a proof of $A \to A$.
}
\label{fig:stack-machine-execution}
\end{figure*}

We formulate proof construction in the Hilbert system as an action sequence executed by a stack machine.
A Hilbert proof is usually represented as a proof tree built from axioms and Modus Ponens.
For a learning algorithm, however, it is more natural to represent proof construction as a sequential decision process.
We therefore flatten proof trees into sequences of stack operations.

Formally, for a target formula \(g\), we define the stack-machine decision process as a deterministic, goal-conditioned decision process given by the tuple \((\mathcal{S}, \mathcal{A}, s_0, P)\), where \(\mathcal{S}\) is the set of finite stacks of formulas, \(\mathcal{A}=\{\Aone, \Atwo, \Athree, \MP\}\) is the action space, and the initial state \(s_0\) is the empty stack.
The transition rule \(P\) is deterministic.
Each axiom action pushes the corresponding Hilbert axiom schema onto the stack.
For the \MP{} action, we use Hindley--Milner unification over formula metavariables~\citep{Hindley1969,Milner1978,DamasMilner1982}.
If the top formula can be unified with an implication \(X \to Y\) and the formula immediately below it can be unified with \(X\), then the \MP{} action is legal and, when applied, pops both formulas and pushes the resulting consequent \(Y\).
The process succeeds for goal \(g\) when the stack consists of the single formula \(g\).
We do not specify a scalar reward function in this formulation, because the learning signals in our method are constructed from self-discovered theorems rather than externally defined rewards.

Figure~\ref{fig:stack-machine-execution} illustrates this view by showing the proof tree for $A \to A$ as a stack-machine execution.
In this example, two \Aone{} actions, one \Atwo{} action, and two Modus Ponens actions produce the theorem $A \to A$ on the stack.
Following this stack-machine decision process, Hilbert-style theorem proving can be treated as a sequential decision problem in which the agent chooses which axiom to introduce and when to apply Modus Ponens.

\section{Self-Supervised Theorem Discovery}
In this section, we describe how the proposed algorithm operates on the stack-machine decision process formulated in Section~\ref{sec:stack-machine-formulation} and how it reuses discovered theorems in subsequent proof search.
The central idea is to construct learning signals from the agent's own search trajectories, rather than relying on external proof data or human-provided lemmas.
Specifically, the policy is learned from self-discovered theorems through goal-conditioned supervised learning.
In addition, the algorithm extracts theorems that are difficult to reprove and highly general, and adds them as new actions so that they can be reused as lemmas in subsequent search.
We describe these components in the following subsections.

\subsection{Learning a Goal-Conditioned Policy from Self-Discovered Theorems}

The proposed method learns a goal-conditioned policy $\pi_\theta(A|S,g)$, where $S$ is the current proof stack, $g$ is the target theorem, and $A$ is a stack-machine action.
Here, $A$ either introduces \Aone, \Atwo, or \Athree, or applies \MP.
The policy therefore predicts which action should be taken next to prove the goal formula $g$ from the current stack $S$.

During search, any state whose stack contains a single formula yields a reached theorem.
If the state $S_t$ contains only one formula $f_t$ at time $t$, we call $S_t$ a single-formula stack and treat $f_t$ as a theorem reached in that episode, regardless of whether it matches the original goal formula $g$.
Thus, even a failed attempt can provide a learning signal when a single-formula stack appears.

\begin{figure*}[t]
\centering
\includegraphics[width=\textwidth]{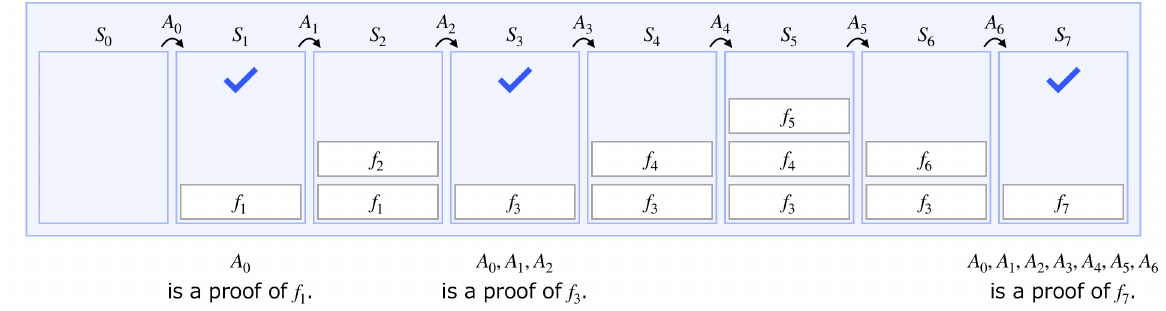}
\caption{An example proof-search episode used to generate goals and training data. Marked states are single-formula stacks, and the corresponding formulas $f_1$, $f_3$, and $f_7$ are added to the goal buffer $\mathcal{G}$. For each reached theorem, we use all preceding state-action prefixes as training examples:
$\mathcal{D}_{\mathrm{train}}\leftarrow
\mathcal{D}_{\mathrm{train}}\cup
\left(\cup_{t\in\{1,3,7\}}\{(S_{t'},A_{t'},f_t)\mid 0\le t'<t\}\right)$
.}
\label{fig:theorem-relabeling}
\end{figure*}

Figure~\ref{fig:theorem-relabeling} illustrates this construction.
The marked states $S_1$, $S_3$, and $S_7$ contain $f_1$, $f_3$, and $f_7$, and the corresponding action prefixes are proofs of those formulas.
These reached theorems are then used in two ways, as future goals and as supervised training data.

We maintain a goal buffer $\mathcal{G}$ containing theorems discovered during search.
Training begins by bootstrapping $\mathcal{G}$ with random rollouts.
Whenever search produces new reached theorems, they are added to $\mathcal{G}$, and later iterations sample a goal formula $g$ uniformly from $\mathcal{G}$ before rolling out the current policy $\pi_\theta(\cdot|S,g)$.

Reached theorems also define supervised targets for the current policy update.
When $f_t$ is reached at time $t$, the prefix up to $t$ is treated as a proof of $f_t$, and for each $0 \le t' < t$ we add $(S_{t'},A_{t'},f_t)$ to $\mathcal{D}_{\mathrm{train}}$.
The policy $\pi_\theta(A|S,g)$ is then updated by cross-entropy on $\mathcal{D}_{\mathrm{train}}$, increasing the likelihood of $\pi_\theta(A_{t'}|S_{t'},f_t)$.

Algorithm~\ref{alg:policy-learning} summarizes this procedure for a fixed action space.
It makes explicit that each reached theorem is inserted into two objects with different roles.
The training set $\mathcal{D}_{\mathrm{train}}$ stores proof-labeled examples for the current policy update and is reset at each iteration.
In contrast, the goal buffer $\mathcal{G}$ is maintained across iterations, so the agent gradually expands the set of goals available for later search.
This separation keeps policy updates local to the current iteration while preserving discovered theorems as persistent future goals.

\begin{algorithm}[tb]
\caption{Policy Learning in a Fixed Action Space}
\label{alg:policy-learning}
\begin{algorithmic}[1]
\STATE {\bfseries Input:} action space $\mathcal{A}$.
\STATE Initialize a new policy $\pi_\theta(A|S,g)$ for $\mathcal{A}$.
\STATE Bootstrap the goal buffer $\mathcal{G}$ with random rollouts.
\FOR{each iteration}
\STATE Initialize current training set $\mathcal{D}_{\mathrm{train}}\leftarrow\emptyset$.
\FOR{each episode}
\STATE Sample a goal formula $g$ uniformly from $\mathcal{G}$.
\STATE Roll out $\pi_\theta(\cdot|S,g)$ for $T$ steps.
\FOR{each reached theorem $f_t$ at time $t$}
\STATE Add $f_t$ to $\mathcal{G}$.
\STATE Add proof-prefix examples for $f_t$ to $\mathcal{D}_{\mathrm{train}}$.
\ENDFOR
\ENDFOR
\STATE Update $\pi_\theta$ by cross-entropy on $\mathcal{D}_{\mathrm{train}}$.
\ENDFOR
\STATE {\bfseries Return:} goal buffer $\mathcal{G}$.
\end{algorithmic}
\end{algorithm}

Our method shares with Hindsight Experience Replay (HER) and goal-conditioned supervised learning (GCSL) the idea of reusing what was actually reached in past trajectories as new goals \citep{Andrychowicz2017HER,Ghosh2021GCSL}.
The difference is that the relabeled goals are not environment states, but proved theorems obtained as single-formula stacks in the Hilbert system.
Thus, formulas encountered during exploration can be reused as valid theorem-proving targets in subsequent search.

\subsection{Extracting Useful Theorems}
We extract useful theorems from the discovered theorem set using two criteria, generality and reprovability.
The goal buffer $\mathcal{G}$ constructed in the previous section serves as the candidate set.
Generality removes overly specialized theorems, while reprovability prioritizes theorems that have already been discovered but cannot be reliably reproved by the current policy.
Adding such theorems as lemmas can expand the set of theorems reached in subsequent proof search.

The first criterion uses the generality order induced by substituting formulas for variables to remove overly specialized theorems.
For example, the theorem $A \to A$ is more general than the theorem $(A \to B) \to (A \to B)$, because substituting the formula $A \to B$ for $A$ in the former yields the latter, whereas no variable substitution in the latter yields the former.
We keep only the theorems that are not specializations of other candidates.
Formally, we define a partial order so that more general theorems are smaller, and retain the minimal elements under this order.
This filtering removes redundant specialized theorems while retaining more general candidates.

The second criterion ranks the remaining candidates by estimating how reliably the current policy can reprove them.
For each candidate theorem $g \in \mathcal{G}$, we record the number $n_g$ of times $g$ was sampled as a proof goal and the number $m_g$ of those episodes in which $g$ was proved.
A theorem with low reprovability has already been discovered, but the current policy cannot reliably prove it again.
Making such a theorem available as a lemma can compress a difficult partial proof into a single lemma call.
To estimate reprovability, we use the Bayesian posterior mean $\hat{p}_g=(m_g+1)/(n_g+2)$, which corresponds to a Beta$(1,1)$ prior, rather than the raw empirical success rate.
This smoothing prevents the ranking from over-prioritizing theorems that fail after only a few samples and helps identify theorems with persistently low reprovability.

Finally, generality filtering and reprovability ranking are combined to select useful theorems from $\mathcal{G}$.
The method first keeps only the theorems that are not specializations of other candidates, thereby removing redundant specializations.
It then sorts the remaining candidates so that lower values of the reprovability posterior mean $\hat{p}_g$ receive higher priority.
Finally, a fixed number of the highest-priority theorems are selected from this ranking and regarded as useful theorems.

\begin{figure*}[t]
\centering
\begin{minipage}[t]{0.48\textwidth}
\centering
\includegraphics[width=\linewidth]{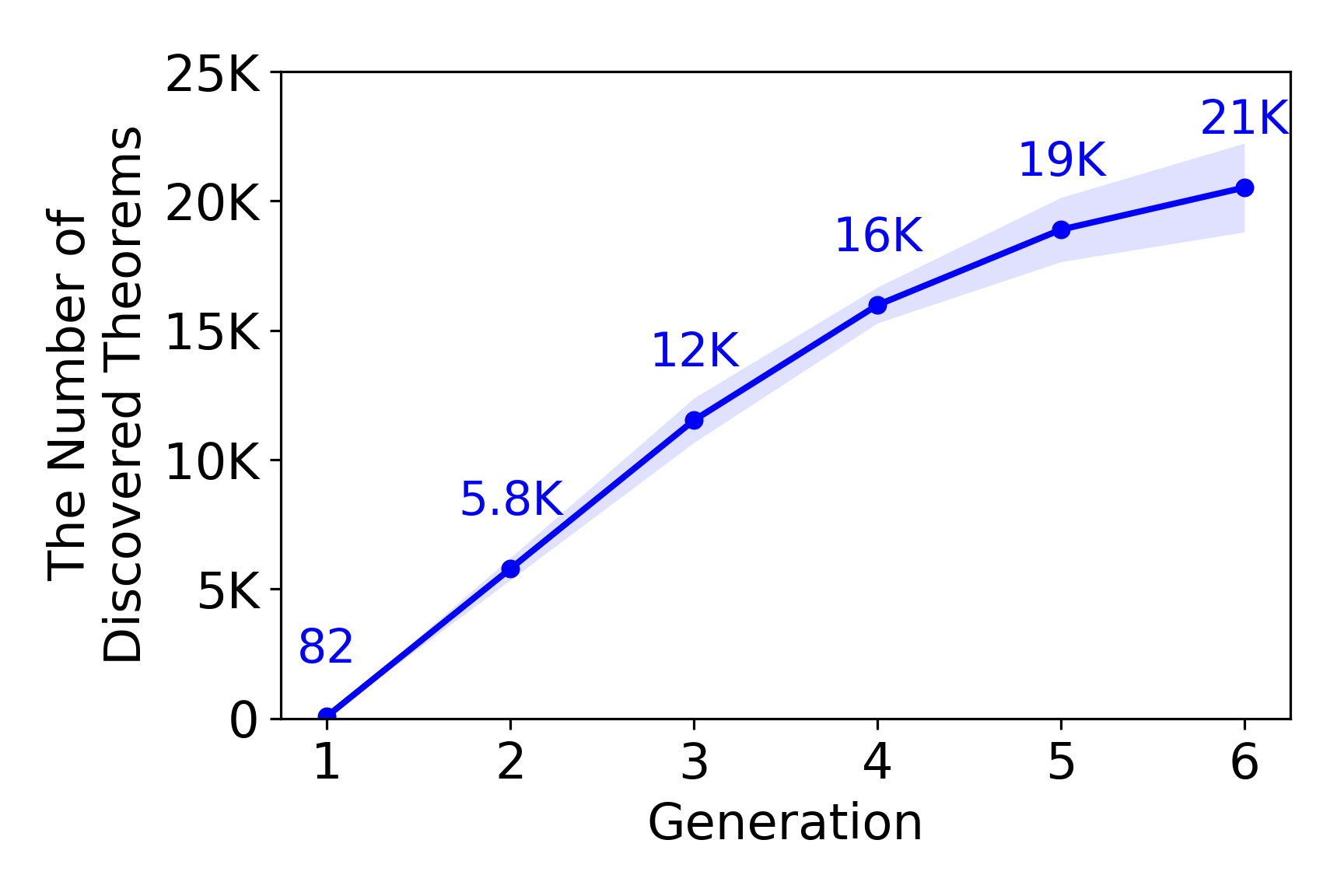}
\caption{Number of discovered theorems in the Hilbert axiom system. The count increases over generations, indicating that the agent can grow a theorem library within the Hilbert axiom system.}
\label{fig:selfplay-found-goals}
\end{minipage}\hfill
\begin{minipage}[t]{0.48\textwidth}
\centering
\includegraphics[width=\linewidth]{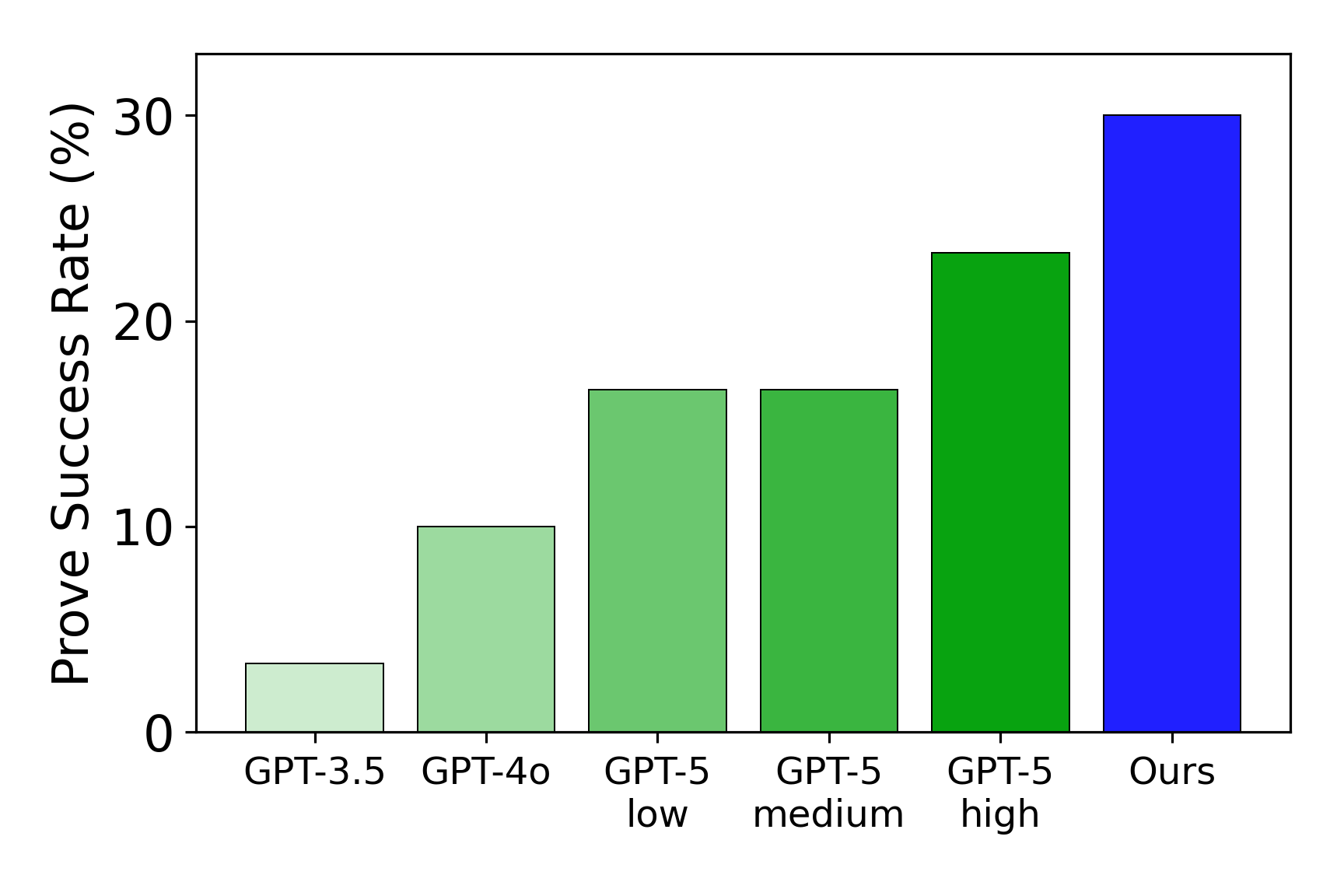}
\caption{Proof success rates on human-written benchmark problems. Our agent finds Hilbert-system proofs for 30\% of them, while GPT baselines indicate the difficulty of the task.}
\label{fig:human-benchmark-performance}
\end{minipage}
\end{figure*}

\subsection{Reusing Useful Theorems}
The proposed method reuses extracted theorems by adding them to the action space of the stack-machine decision process.
The original action space consists of four actions: pushing an instance of \Aone, \Atwo, or \Athree{} onto the stack, and applying the inference rule \MP.
For each extracted theorem, we add one library action, denoted by $\LibAct{1},\LibAct{2},\ldots,\LibAct{N}$.
When one of these theorem actions is selected, the corresponding theorem is pushed onto the stack instead of an axiom.
The resulting stack-machine decision process allows the agent to reuse theorems discovered in earlier generations as lemmas in later proof search.

Algorithm~\ref{alg:overall} summarizes how the action space is expanded across generations.
The algorithm alternates between policy learning under a fixed action space and action-space expansion using extracted theorems.
In each generation, we train a new policy network for the current action space, because adding library actions changes the action space of the stack-machine decision process.
During policy learning, reached theorems accumulate in the goal buffer $\mathcal{G}$ through multiple iterations of proof search and supervised updates.
At the end of the generation, useful theorems are extracted from $\mathcal{G}$, and the corresponding library actions are added to the action space $\mathcal{A}$.
This process turns theorems discovered in one generation into reusable action primitives for later generations.
By repeating this process, the agent gradually expands its action space and grows a library of useful theorems.

\begin{algorithm}[htb]
\caption{Multi-Generation Library Expansion}
\label{alg:overall}
\begin{algorithmic}[1]
\STATE Initialize action space $\mathcal{A}\leftarrow\{\Aone,\Atwo,\Athree,\MP\}$.
\FOR{each generation}
\STATE Give $\mathcal{A}$ to Algorithm~\ref{alg:policy-learning} and obtain theorem buffer $\mathcal{G}$.
\STATE Extract useful theorems $\{\LibAct{1},\ldots,\LibAct{N}\}$ from $\mathcal{G}$.
\STATE Expand $\mathcal{A}\leftarrow\mathcal{A}\cup\{\LibAct{1},\ldots,\LibAct{N}\}$.
\ENDFOR
\end{algorithmic}
\end{algorithm}

\section{Experiments}
\subsection{Experimental Setup}
\begin{table*}[t]
\caption{Qualitative examples in which GPT-5-high failed without prompt lemmas but succeeded when supplied with extracted theorem actions. The last column lists the theorem actions used in each successful proof.}
\label{tab:qualitative-goals}
\centering
\begin{tabular}{@{}c@{\hspace{0.8em}}ll@{}}
\toprule
Example & Goal formula & Used theorem actions \\
\midrule
1 & $A \to (\neg A \to B)$ & \LibAct{1}, \LibAct{2} \\
2 & $(A \to B) \to ((B \to C) \to (A \to C))$ & \LibAct{3}, \LibAct{5} \\
3 & $\neg((A \to B) \to \neg(B \to A)) \to (B \to A)$ & \LibAct{1}, \LibAct{2}, \LibAct{3}, \LibAct{4} \\
\bottomrule
\end{tabular}
\end{table*}

\begin{table*}[t]
\caption{Extracted theorem actions appearing in the successful qualitative proofs in Table~\ref{tab:qualitative-goals}. Each action label denotes a discovered theorem supplied to GPT-5-high as a prompt lemma.}
\label{tab:qualitative-actions}
\centering
\begin{tabular}{@{}cl@{}}
\toprule
Action & Theorem formula \\
\midrule
\LibAct{1} & $\neg A \to (A \to B)$ \\
\LibAct{2} & $(A \to (B \to C)) \to (D \to ((A \to B) \to (A \to C)))$ \\
\LibAct{3} & $(A \to B) \to ((C \to A) \to (C \to B))$ \\
\LibAct{4} & $(\neg A \to \neg(B \to (C \to B))) \to A$ \\
\LibAct{5} & $((A \to (B \to C)) \to (((A \to B) \to (A \to C)) \to D)) \to ((A \to (B \to C)) \to D)$ \\
\bottomrule
\end{tabular}
\end{table*}

We run theorem discovery and library construction in the Hilbert axiom system shown in Figure~\ref{fig:hilbert-system}.
The initial action space consists of \Aone, \Atwo, \Athree, and \MP{}.
Thus, before any library theorem is extracted, the agent can only push instances of the three axioms onto the stack or apply Modus Ponens.
We repeat proof search and library extraction over six generations.
The numbers of extracted theorems added to the library are $20,10,5,2,1,0$ from Generation 1 to Generation 6, which yields a final library of $38$ theorem actions.

For benchmark evaluation, we use the $30$ human-written propositional-logic problems derived from Kleene's textbook \citep{Kleene1952} and curated by \citet{Poesia2024Minimo}.
This benchmark provides a human-written test set that is independent of our theorem-discovery process.
Each LLM baseline is evaluated once per problem with fixed decoding settings, so the reported success rates reflect single-shot proof generation rather than best-of-$k$ sampling.
A problem is counted as solved only when the generated action sequence is accepted by a stack-machine proof checker implementing \Aone--\Athree, theorem actions, and Modus Ponens with the same unification rule as in Section~\ref{sec:stack-machine-formulation}.
In the lemma-augmented setting, each theorem action pushes the corresponding theorem from the extracted library onto the stack.
Since each extracted theorem is stored with its primitive axiom/\MP{} proof trajectory, lemma-augmented proofs can in principle be expanded into axiom-only Hilbert proofs.

\subsection{Quantitative Evaluation of Theorem Discovery}
We first evaluate whether the proposed method can grow the goal buffer $\mathcal{G}$ starting only from the axioms and inference rule of the Hilbert axiom system defined in Section~\ref{sec:preliminaries}.

Figure~\ref{fig:selfplay-found-goals} shows that the number of discovered theorems increases over generations.
This result demonstrates that our approach accumulates found theorems and reuses them to derive further theorems within the formal axiomatic system.

We next evaluate coverage on the human-written benchmark problems.
Discovering many theorems alone does not guarantee that they are meaningful from a human perspective.
This benchmark consists of classical-logic problems.
As shown in Figure~\ref{fig:human-benchmark-performance}, the discovered theorem set covers $30\%$ of these benchmark problems.
This result suggests that the agent does not merely generate many theorems, but also discovers theorems that are meaningful from a human perspective.

We further evaluate the difficulty of solving these benchmark problems in the Hilbert system using LLM baselines.
Figure~\ref{fig:human-benchmark-performance} shows the fraction of benchmark problems for which each GPT baseline \citep{OpenAI2023GPT35API,OpenAI2024GPT4oSystemCard,OpenAI2025GPT5SystemCard} finds a Hilbert-system proof path.
For GPT-5, low, medium, and high denote the reasoning effort used at inference time.
Even GPT-5-high, the strongest GPT baseline in our evaluation, reaches only about $23\%$ accuracy, which indicates that constructing Hilbert-system proofs for these benchmark theorems is nontrivial.
Here, we do not aim to claim that the proposed method generally outperforms state-of-the-art LLMs.
Rather, these results show that the benchmark problems are difficult to solve in the Hilbert system and indicate that our approach can discover nontrivial proofs.

\subsection{Quantitative Evaluation of Extracted Theorems}
We evaluate whether the extracted theorems also function as external knowledge for LLM proof search.
Specifically, we provide the extracted theorems as prompt lemmas and compare the proof success rate of GPT baselines on human-written benchmark problems.
This evaluation tests whether theorems obtained by our proof-search agent can transfer to LLMs, which rely on a different reasoning architecture.

\begin{figure}[t]
\centering
\includegraphics[width=\linewidth]{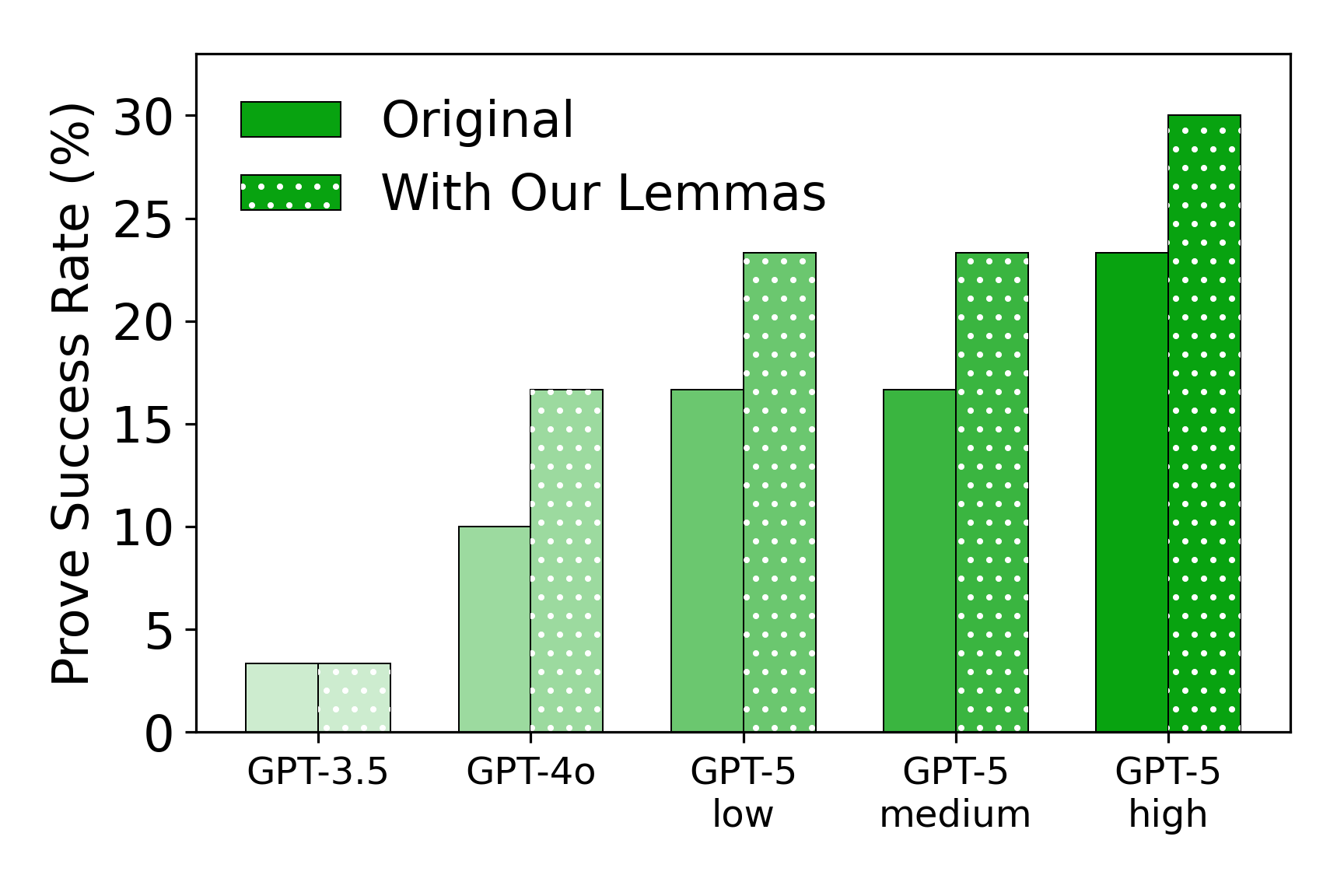}
\caption{Effect of extracted theorems as prompt lemmas for LLM proof search. Providing extracted theorems improves proof success rates for GPT-4o and GPT-5 variants.}
\label{fig:llm-lemma-transfer}
\end{figure}

Figure~\ref{fig:llm-lemma-transfer} compares the proof success rate with and without the extracted theorems provided as prompt lemmas.
For GPT-4o and the GPT-5 variants, proof success rates improve when the extracted theorems are provided as prompt lemmas.
This result shows that the extracted theorems can help LLMs find proofs in the Hilbert system.
Overall, these results show that theorems extracted through our approach function as transferable knowledge for LLM-based theorem proving.

\subsection{Qualitative Analysis of Extracted Theorems}
\label{sec:qualitative-extracted-theorems}
Finally, we examine concrete examples of how extracted theorems assist LLM proof construction.
Table~\ref{tab:qualitative-goals} summarizes three goals for which GPT-5-high failed to find a proof without prompt lemmas but succeeded when the extracted theorem library was supplied.
Table~\ref{tab:qualitative-actions} lists the extracted theorem actions that appear in these successful proofs.

The first example is especially informative because the two lemmas used in the proof have different roles.
With the extracted theorems provided as prompt lemmas, GPT-5-high constructed the following Hilbert-system proof action sequence:
\begin{equation*}
\Aone,\LibAct{1},\LibAct{2},\MP,\Atwo,\MP,\MP.
\end{equation*}

In this proof, \LibAct{1} corresponds to the principle of explosion, deriving an arbitrary $B$ from $\neg A$ and $A$.
Thus, the usefulness of \LibAct{1} is easy to interpret from a human perspective.
In contrast, \LibAct{2} does not appear among the human-written benchmark problems and was discovered by our approach.
Although its usefulness is not obvious at first glance, \LibAct{2} lifts an implication under an additional context and contributes to proving the first goal in Table~\ref{tab:qualitative-goals} when combined with \LibAct{1}.
This contrast shows that usefulness is not always apparent from human inspection alone.

These examples show that extracted theorems do not merely improve aggregate evaluation metrics, but also play concrete roles in individual proof constructions.
In particular, not only human-interpretable theorems such as \LibAct{1} and \LibAct{3}, but also theorems whose usefulness is not obvious at first glance, such as \LibAct{2}, \LibAct{4}, and \LibAct{5}, function as effective supporting knowledge when an LLM constructs Hilbert-system proofs.
More broadly, this suggests that theorem discovery without human prior knowledge may allow AI systems to surface useful lemmas that humans did not explicitly provide.

\section{Related Work}
\paragraph{Learning-guided formal theorem proving.}
Prior work on learning-guided formal theorem proving has studied reinforcement-learning proof search, tactic prediction, proof generation, and LLM-based proving in Lean \citep{Kaliszyk2018RLTheoremProving,Han2022PACT,Polu2022StatementCurriculum,Yang2023LeanDojo,Xin2024DeepSeekProver,Zimmer2025Bourbaki,Hubert2026AlphaProof}.
These methods build strong provers using existing formal libraries or proof data.
In contrast, we ask how far theorem discovery and reuse can go using only the three axioms and one inference rule of a Hilbert axiom system.

\paragraph{Self-play and intrinsic theorem discovery.}
Another line reduces dependence on human proof examples by learning proof-search or conjecturing through self-play or intrinsic motivation \citep{Wu2021TacticZero,Lample2022HyperTree,Poesia2024Minimo,Laurent2022RefineSearch,Dong2025STP,Zhao2025AbsoluteZero}.
\citet{Kasriel2025Usefulness} is especially close because it discovers useful theorems from axioms, reuses them for further search, and evaluates external usefulness by asking an LLM judge whether they appear useful.
Complementing this direction, we ask whether self-supervised theorem-discovery methods can find theorems that not only appear useful, but also actually improve LLM-based proof search when supplied as lemmas.

\paragraph{Goal-conditioned and hindsight learning.}
Our learning rule relates to Hindsight Experience Replay and goal-conditioned supervised learning, which reuse reached states as goals \citep{Andrychowicz2017HER,Ghosh2021GCSL}.
In our setting, the reached objects are theorems, not environment states, and some are reused as both training goals and action primitives in subsequent stack-machine decision processes.
Thus, a proof prefix reaching a formula becomes hindsight supervision for proving it.
This connects goal relabeling with theorem-proving library learning.

\paragraph{Lemma discovery and library learning.}
Lemma discovery and library learning are important directions for accelerating proof search \citep{Yang2023LeanDojo,Wang2024LEGOProver}.
In contrast to methods that rely on existing formal libraries, human-provided lemmas, or LLM-based proving pipelines, our agent discovers useful theorems through proof search using only the axioms and inference rule.

\paragraph{Curry--Howard motivation.}
Our stack-machine view is also motivated by the Curry--Howard correspondence, which relates propositions to types and proofs to programs \citep{Howard1980,SorensenUrzyczyn2006}.
There, implication resembles a function type and \MP{} resembles function application.
This is not central to our claims, but explains why Hilbert proofs can be flattened into action sequences that push axioms or lemmas and compose them on a stack.
Our \MP{} implementation follows this intuition by unifying an implication antecedent with another stack formula, as in Hindley--Milner type inference \citep{Hindley1969,Milner1978,DamasMilner1982}.

\section{Conclusions}
We studied whether useful theorems can be discovered within a formal axiomatic system using only its primitive rules.
We formulated Hilbert-system propositional theorem proving as a stack-machine decision process and proposed a learning algorithm that grows a theorem library through proof search.
The algorithm treats reached theorems as future proof goals and extracts useful theorems for reuse as lemmas.
Experiments show that the agent discovers many theorems over multiple generations in the Hilbert axiom system and finds proofs for a meaningful fraction of the human-written benchmark problems.
We also demonstrate that the extracted theorems improve LLM proof performance when provided as prompt lemmas.

A limitation of this study is that our experiments are conducted in a single axiom system.
However, the proposed framework is not specific to this axiom system and should be applicable to multiple axiom systems; evaluating this broader applicability is a natural next step.
Furthermore, it also remains an important direction for future work to test whether the same approach can scale to richer formal systems with broader applications, such as first-order logic, set theory, and more general mathematical domains.

While we have focused on Hilbert-style propositional theorem proving, our experiments provide evidence that useful theorems can emerge from proof search without relying on human-provided theorem libraries.
More broadly, these findings suggest a path toward self-evolving AI systems for mathematics: systems that autonomously expand their formal knowledge through proof search while keeping each discovery formally verifiable.

\section*{Acknowledgements}
This work was partially supported by JST Moonshot R\&D Grant Number JPMJPS2011, CREST Grant Number JPMJCR2015 and Basic Research Grant (Super AI) of Institute for AI and Beyond of the University of Tokyo.
Kazuki Ota was supported by JST SPRING, Grant Number JPMJSP2108.
Takayuki Osa was supported by JSPS KAKENHI Grant Number JP25K03176.

\bibliography{references}
\bibliographystyle{icml2026}

\clearpage
\appendix
\onecolumn
\section{Experimental Details}

In this section, we explain implementation details for theorem-discovery training and library construction.
Each generation uses $25$ training iterations.
In each iteration, the agent generates $8192$ episodes, and each episode has a maximum length of $7$ steps.
Therefore, each generation uses $25 \times 8192 \times 7$ simulator evaluations.
Episodes are processed in batches of $2048$ episodes.

The policy is parameterized by a Transformer network.
The model uses embedding dimension $128$, four Transformer blocks, four attention heads, feed-forward dimension $512$, dropout rate $0.1$, and learned positional encodings.
The policy is trained with cross-entropy loss under a legal-action mask.
We use Adam with learning rate $0.001$, batch size $512$, and one training epoch per iteration.

The policy observation is a fixed-length sequence of $1024$ integer tokens.
It consists of the sampled goal formula followed by the current proof stack, separated by a special field-separator token, and padded to a fixed length.
Each formula is tokenized at the symbol level in reverse Polish notation, with tokens for implication, falsity, variables, and formula separators.
Thus, a training example for $\pi_\theta(A|S,g)$ presents the goal formula $g$ and the current stack $S$ to the Transformer.

Across generations, the action space consists of the three axiom actions, the \MP{} action, and the library theorem actions available at that generation.
The legal-action mask is computed from the current stack, the remaining horizon, and the current action space.
It is used to mask illegal actions during sampling and in the cross-entropy loss.
When new theorems are extracted and added to the library, the action space changes accordingly.
We therefore initialize and train a new policy network for the updated action space at each generation.

\section{Prompts for LLM Evaluation}

We show the prompt template used for LLM evaluation.
In the prompt, the axioms corresponding to \Aone{}, \Atwo{}, and \Athree{} in the main text are denoted by `A01', `A02', and `A03', respectively, and Modus Ponens is denoted by `MP'.
When discovered theorems are provided as lemmas, they are listed as `L01', `L02', and so on.
For the baseline without lemmas, this lemma list is empty, and the LLM is asked to construct a proof using only `A01', `A02', `A03', and `MP'.
The final answer is requested as an RPN sequence in which actions are separated by spaces.

\begin{lstlisting}
# Rules
Prove the given proposition in the Hilbert system (with axioms A01, A02, A03 and the inference rule Modus Ponens).

## Axioms:
A01: A -> (B -> A)
A02: (A -> (B -> C)) -> ((A -> B) -> (A -> C))
A03: (!A -> !B) -> (B -> A)

## Inference Rule (MP):
In the stack, the top element is regarded as X->Y, and the one below it as X. Unify the antecedents of X and X->Y as in Hindley-Milner type inference, and as a result derive the consequent Y, which is then pushed onto the stack. For example, if the top of the stack is (A->(B->C))->((A->B)->(A->C)) and the next one is A->(B->A), then A->(B->C) and A->(B->A) are unified with A=C, yielding the result (A->B)->(A->A).

## Lemmas: 
You can use the following lemmas if you need.
L01: (A -> B) -> ((C -> A) -> (C -> B))
L02: (A -> (B -> C)) -> (B -> (A -> C))

# Output Format
The proof must ultimately be represented as a combinator expression in Reverse Polish Notation (RPN). In RPN, A01, A02, A03 each push an instance of the corresponding axiom onto the stack, L01, L02, ... do the same for lemmas, and MP represents Modus Ponens according to the above rule. You may include explanations, but enclose the final proof expression between <final_answer> and </final_answer>.

# Example Problem
Target proposition: A -> A
Reference proof steps:
1. A -> (B -> A) (Axiom A01)
2. A -> (B -> A) (Axiom A01)
3. (A -> (B -> C)) -> ((A -> B) -> (A -> C)) (Axiom A02)
4. (A -> B) -> (A -> A) (Apply MP to 3 and 2)
5. A -> A (Apply MP to 4 and 1)
In this case, the combinator expression in RPN is A01 A01 A02 MP MP. Final output example is the following: 
<final_answer>A01 A01 A02 MP MP</final_answer>

# Problem
Prove the following proposition:
(A -> B) -> ((B -> C) -> (A -> C))
\end{lstlisting}

\end{document}